\documentclass{article}

\usepackage[final]{tccml_neurips_2020}

\usepackage[utf8]{inputenc} 
\usepackage[T1]{fontenc}    
\usepackage{hyperref}       
\usepackage{url}            
\usepackage{float}
\usepackage{booktabs}       
\usepackage{amsfonts}       
\usepackage{amsmath}
\usepackage{amssymb}
\usepackage{nicefrac}       
\usepackage{microtype}      
\usepackage[dvipsnames]{xcolor}

\newcommand\restr[2]{{
  \left.\kern-\nulldelimiterspace 
  #1 
  \vphantom{\big|} 
  \right|_{#2} 
  }}

\usepackage{graphicx}
\usepackage{wrapfig}
\usepackage{subcaption}
\usepackage{tabularx}
\usepackage{float}
\usepackage[title]{appendix}
\usepackage{gensymb}
\usepackage{capt-of}

\title{EarthNet2021: A novel large-scale dataset and challenge for forecasting localized climate impacts.}

\author{%
  Christian Requena-Mesa\textsuperscript{1,2,3,*}
  \And
  Vitus Benson\textsuperscript{1,*}
  \AND
  Joachim Denzler\textsuperscript{2,4} \\
  \And
  Jakob Runge\textsuperscript{3} \\
  \And
  Markus Reichstein\textsuperscript{1,4} \\
  \end{tabular}\hfil\linebreak[0]\hfil%
  \small
    \begin{tabular}[t]{l}
  \textsuperscript{1} Department Biogeochemical Integration, Max-Planck-Institute for Biogeochemistry, Jena \rule{0pt}{12pt}\\
  \textsuperscript{2} Computer Vision Group, University of Jena\\
  \textsuperscript{3} Institute of Data Science, German Aerospace Center (DLR), Jena\\
  \textsuperscript{4} Michael-Stifel-Center Jena for Data-driven and Simulation Science\\
  \textsuperscript{*} Joint first authors. E-Mail: \{crequ, vbenson\}@bgc-jena.mpg.de
  \rule{0pt}{2pt}
}

\begin{document}

\maketitle
\begin{minipage}[t][0pt]{\linewidth}
\begin{abstract}
Climate change is global, yet its concrete impacts can strongly vary between different locations in the same region. Regional climate projections and seasonal weather forecasts currently operate at the mesoscale (\textgreater~$1$~km). For more targeted mitigation and adaptation, modelling impacts to \textless~$100$~m is needed. Yet, the relationship between driving variables and Earth’s surface at such local scales remains unresolved by current physical models and is partly unknown; hence, it is a source of considerable uncertainty. 
Large Earth observation datasets now enable us to create machine learning models capable of translating coarse weather information into high-resolution Earth surface forecasts encompassing localized climate impacts.
Here, we define high-resolution Earth surface forecasting as video prediction of satellite imagery conditional on mesoscale weather forecasts. Video prediction has been tackled with deep learning models.
Developing such models requires analysis-ready datasets. We introduce \textbf{EarthNet2021}, a new, curated dataset containing target spatio-temporal Sentinel~2 satellite imagery at $20$~m resolution, matched with high-resolution topography and mesoscale ($1.28$~km) weather variables.
With over $32000$ samples it is suitable for training deep neural networks. Comparing multiple Earth surface forecasts is not trivial. Hence,  we define the EarthNetScore, a novel ranking criterion for models forecasting Earth surface reflectance. For model intercomparison we frame EarthNet2021 as a challenge with four tracks based on different test sets. These allow evaluation of model validity and robustness as well as model applicability to extreme events and the complete annual vegetation cycle. In addition to forecasting directly observable weather impacts through satellite-derived vegetation indices, capable Earth surface models will enable downstream applications such as crop yield prediction, forest health assessments, coastline management, or biodiversity monitoring. Find data, code, and how to participate at \url{www.earthnet.tech}.
\end{abstract}
\end{minipage}
\newpage

\begin{figure}[t]
    \centering
    \includegraphics[width = \textwidth]{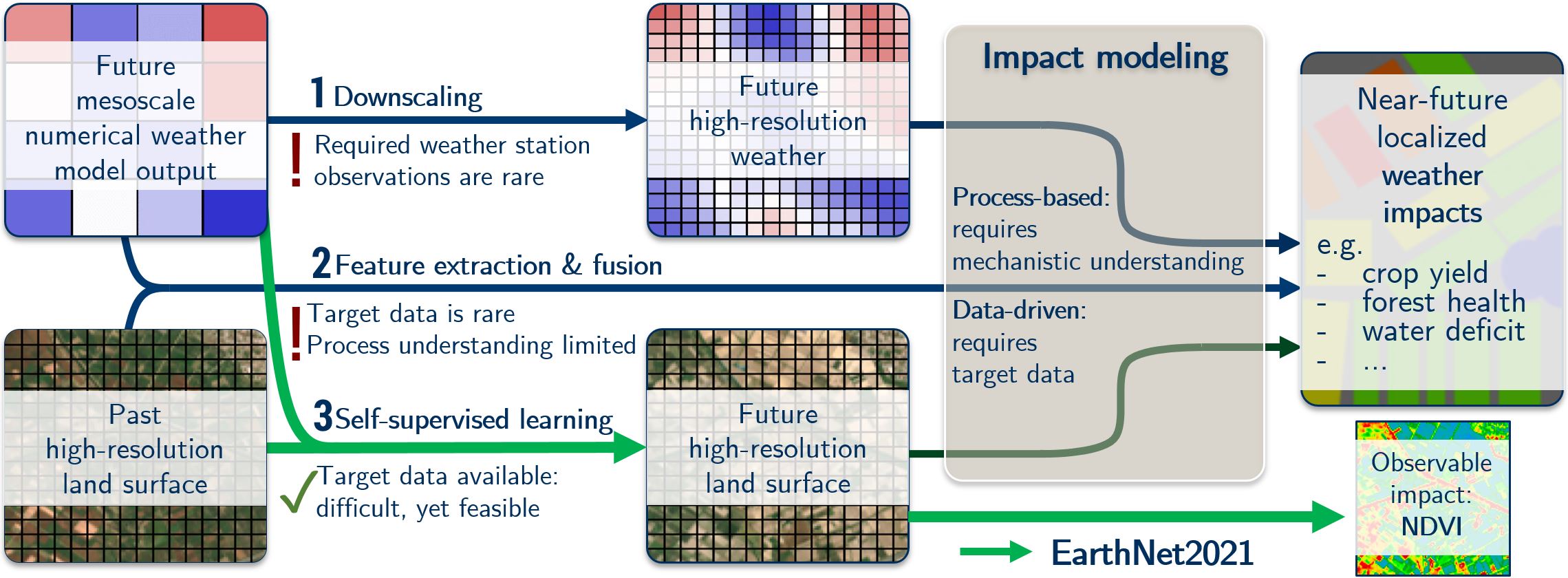}
    \caption{Predicting localized climate impacts can be done in a variety of ways. EarthNet2021 places Earth surface prediction as an intermediate task (green path). This gives directly obtainable impacts (NDVI) and enables further analyses to extract impacts from future Earth surface reflectance (lower dark green path in gray box). Compared to directly modeling any given impacts, where target labels are scarce, large amounts of satellite imagery are available. Hence, localized prediction of climate impacts becomes feasible since self-supervised deep learning can be leveraged.}
    \label{fig:why}
\end{figure}

\section{Motivation}
The terrestrial surface of Earth is home to most of the planet's species and houses human economical and societal systems. As already noticed by Alexander von Humboldt, climate is a key factor shaping vegetation cover and soils on Earth.  Yet, importantly, the impact of the climatic drivers onto the surface is highly modulated by the fine-grained local conditions, such as geomorphology, geological substrate and vegetation and animals themselves. In particular extreme events can have very heterogeneous impacts at the local scale depending on the conditions \citep{kogan1990remote}.  For example, ecosystems next to a river might survive droughts better than those on south-facing slopes. However, the resolution of seasonal weather predictions is not fine enough to deploy effective prevention and mitigation strategies. Machine learning \citep{reichstein2019, rolnick2019} can step in to increase resolution.

Predicting localized climate impacts can be tackled in three main ways (see Fig.~\ref{fig:why}). All approaches make use of seasonal weather forecasts \citep[$2$ -- $6$ months ahead;][]{cantelaube2005seasonal}. The first approach (Fig.~\ref{fig:why}, path 1), aims to reconstruct hyper-resolution weather forecasts for particular geolocations using statistical \citep{boe2006simple, vrac2007statistical} or dynamical \citep{lo2008assessment} downscaling, that is, correlating the past observed weather with past mesoscale model outputs and using the estimated relationship. The downscaled weather can then be used in mechanistic models (e.g. of river discharge) for impact extraction. However, weather downscaling is a difficult task because it requires ground observations from weather stations, which are sparse.

A more direct way (Fig.~\ref{fig:why}, path 2) is to correlate a desired future impact variable with past data. Doing this for tangible impacts such as on crop yield \citep{peng2018} again suffers from a lack of data. An alternative is to forecast impacts obtainable from remote sensing, such as the normalized differenced vegetation index (NDVI), yet this has only been done at coarse resolution \citep{tadesse2010,asoka2015,kraft2019,foley2020}.

Instead, we propose \emph{Earth surface forecasting} as video prediction of satellite imagery with guidance of mesoscale weather projections for forecasting localized weather impacts (Fig.~\ref{fig:why}, path 3). From satellite imagery, we can directly observe NDVI and thus climate impacts on vegetation. Additionally, it can also be used to extract further processed weather impact data products, such as the biodiversity state \citep{fauvel2020}, crop yields \citep{schwalbert2020}, soil moisture \citep{efremova2019}, or ground biomass \citep{ploton2017}. We believe Earth surface forecasting is feasible since numerous studies suggest predicting satellite imagery works under a range of specific conditions \citep{zhu2015, das2016, hong2017, requena2018, lee2019}.

EarthNet2021 aims at providing analysis-ready data and a benchmark challenge for model intercomparison on the broad area of Europe to accelerate model development.

{

\begin{figure}[bt]
\centering
    \includegraphics[width=1\textwidth]{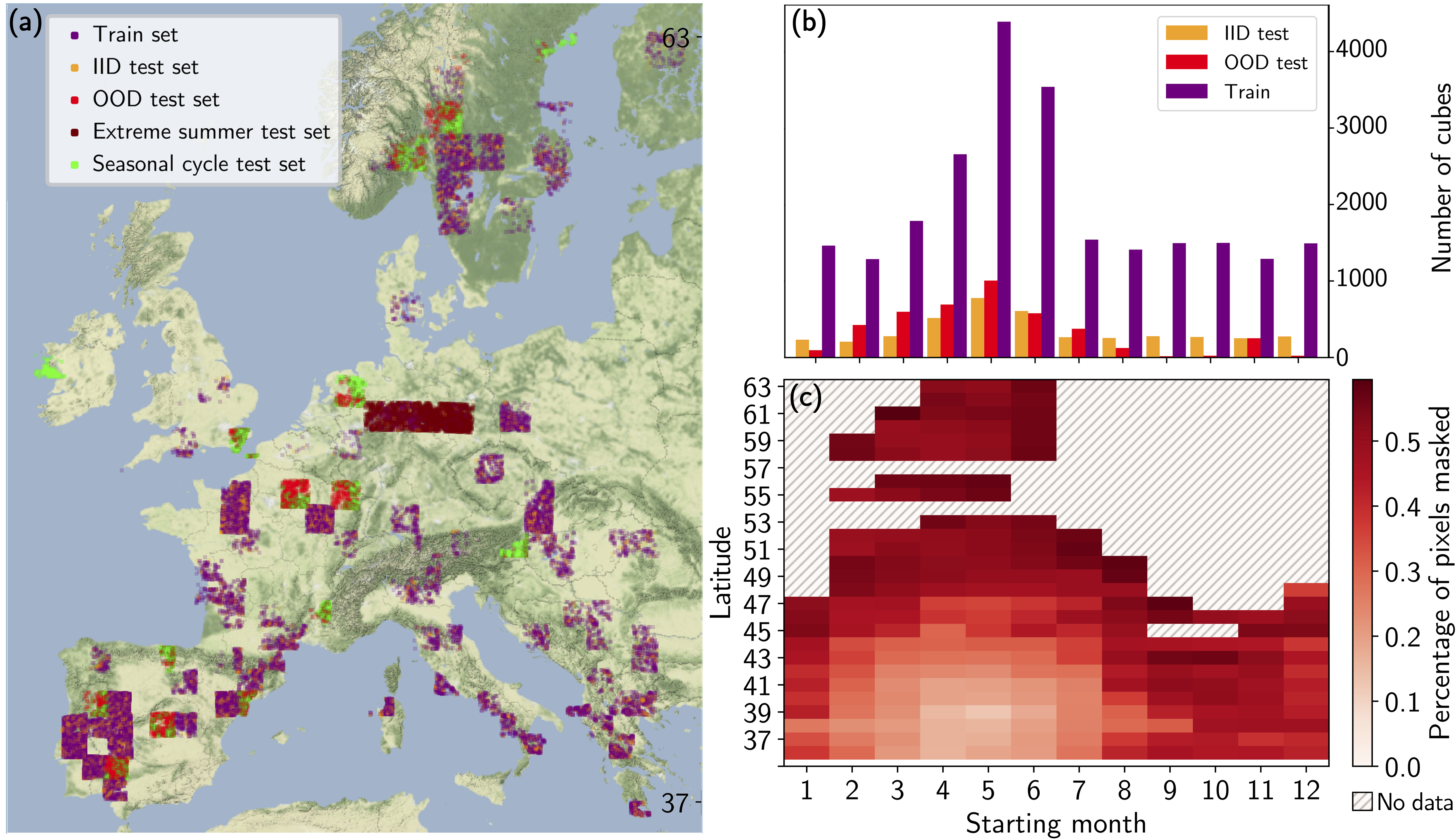}
  \caption{Panel (a) shows the spatial distribution of the multicubes in EarthNet2021. Panel (b) shows the monthly number of multicubes and panel (c) shows the data quality measured by the percentage of masked (mainly cloudy) pixels over both, months and latitude.}
  
  \label{fig:bias}
  %
\end{figure} 
}

\begin{wrapfigure}{R}{0.5\textwidth}
    \centering
    \includegraphics[width=0.5\textwidth]{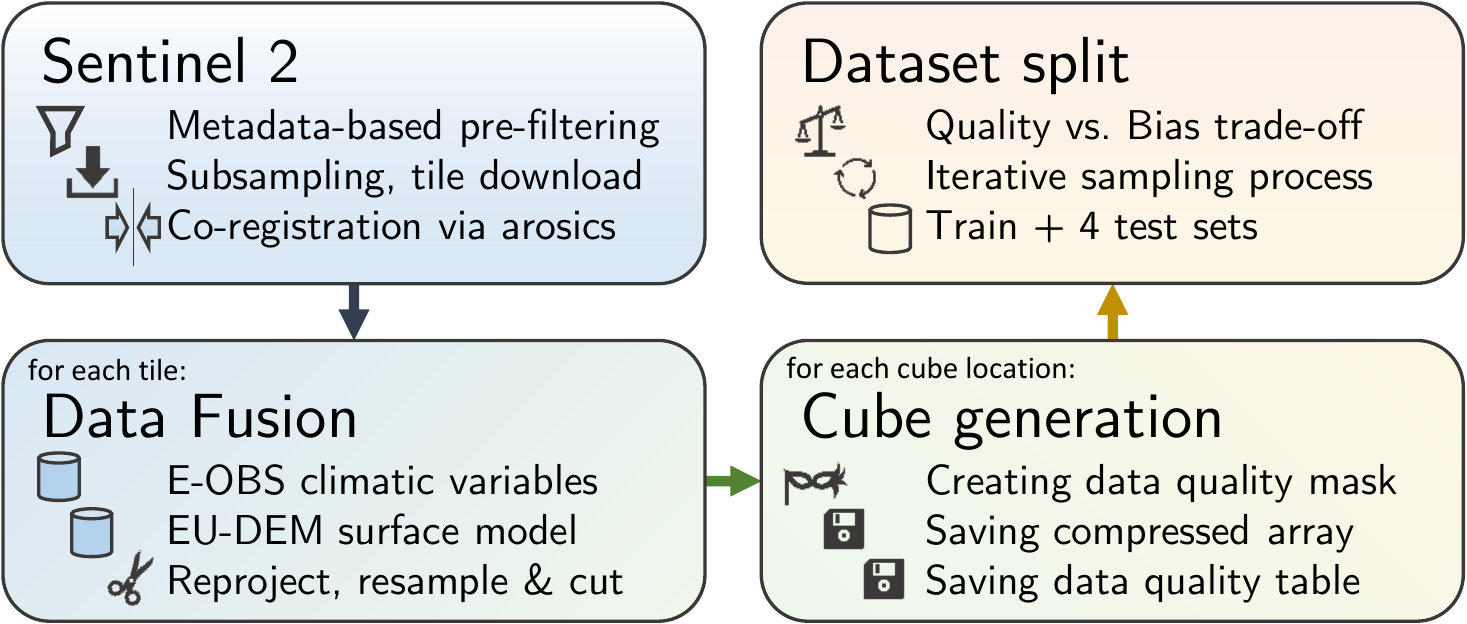}
    \caption{This dataset generation scheme roughly describes how we obtain the analysis-ready EarthNet2021 dataset from raw, public geospatial datasets.}
    \label{fig:datagen}
\end{wrapfigure}

\section{Data}

The EarthNet2021 dataset combines three publicly available EU-funded data sources into spatio-temporal data samples for training deep learning models. Each sample, which we call a data multicube, is based on a timeseries of Sentinel 2 level 2A imagery \citep{louis2016} combined with a timeseries of daily climatic conditions from E-OBS \citep{cornes2018} and the EU-DEM digital surface model \citep{bashfield2011}. Training deep learning models with raw geospatial data is usually not possible and there is need for analysis-ready datasets. An overview of the dataset generation pipeline, turning the raw geospatial data into the analysis-ready EarthNet2021 dataset, is shown in Fig.~\ref{fig:datagen}.

After data processing, EarthNet2021 contains over $30000$ samples, which we call \emph{data multicubes}. A single multicube is visualized in Fig.~\ref{fig:3d}. It contains $30$ 5-daily frames ($128 \times 128$ pixel or $2.56 \times 2.56$ km) of four channels (blue, green, red, near-infrared) of satellite imagery with binary quality masks at high-resolution ($20$ m), $150$ daily frames ($80 \times 80$ pixel or $102.4 \times 102.4$ km) of five dynamic climatic variables (precipitation, sea level pressure, mean, minimum and maximum temperature) at mesoscale resolution ($1.28$ km) and a static digital elevation model at both high- and mesoscale resolution.

The entirety of the multicubes have been split across the training set and various test sets, which are related to different tracks in the EarthNet2021 challenge (sec.~\ref{sec:challenge}). EarthNet2021 is an imbalanced dataset, as during the data generation there is a direct trade-off between high data quality and low selection bias. For example, high-quality (cloud-free) samples are mostly found during summer on the Iberian Peninsula, whereas there are few consecutive weeks without clouds on the British Islands. In Fig.~\ref{fig:bias} we try to make some of the selection bias visible by showing the spatial and temporal distribution of the multicubes in EarthNet2021 across the different sets.

\begin{wrapfigure}[23]{R}{0.5\textwidth}
    \vspace{-40pt}
    \centering
    \includegraphics[width=0.5\textwidth]{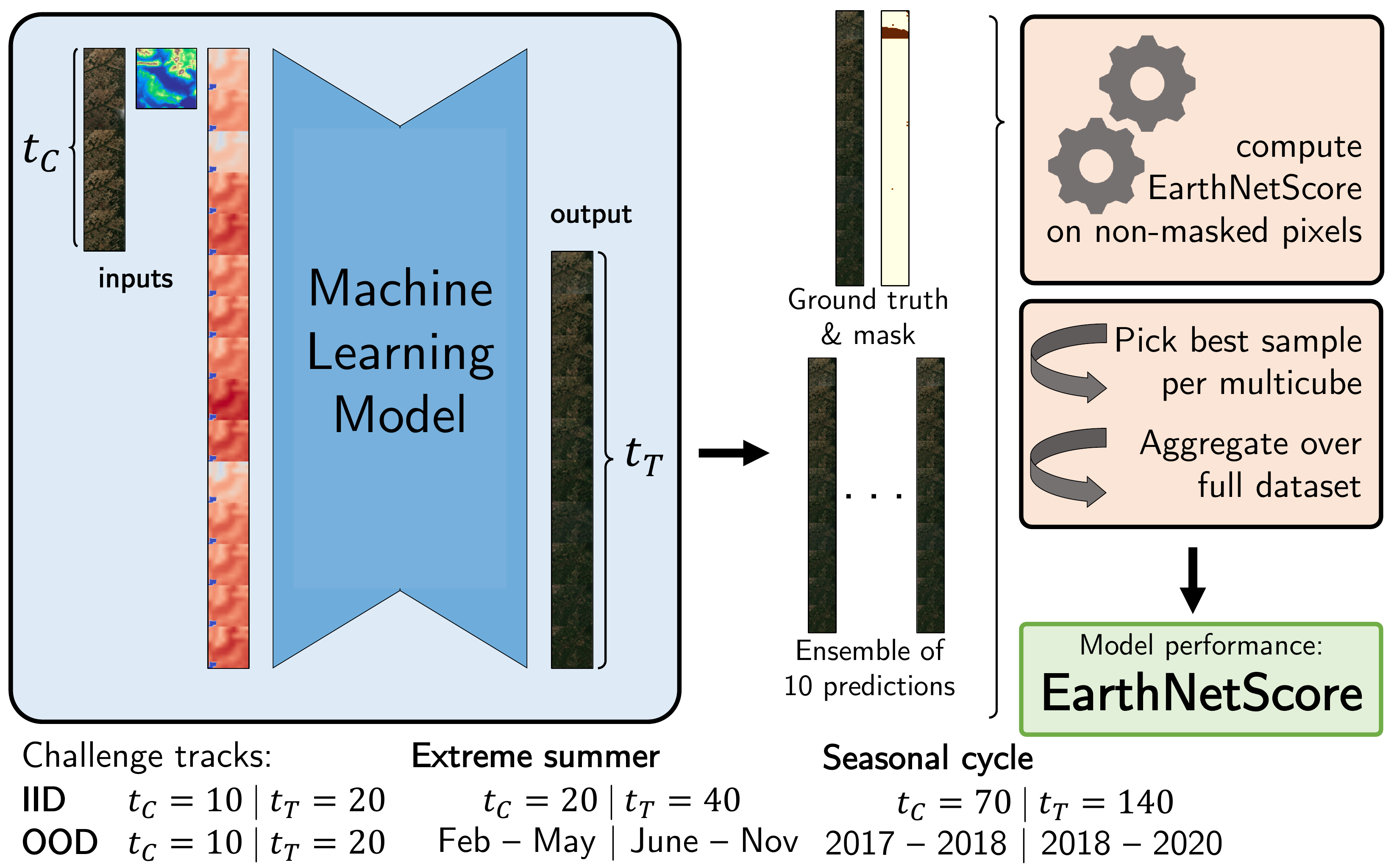}
    \caption{Evaluation pipeline for models on EarthNet2021. For predicting the $t_T$ target frames, a model can use satellite images from the $t_C$ context frames, the static DEM and mesoscale climatic variables including those from the target time steps.}
    \label{fig:eval}
    \vspace*{10pt}
    
    {\small
    \setlength{\tabcolsep}{2pt}
	\centering
	\begin{tabularx}{0.5\textwidth}{Xcccccc}
		\toprule
		Test set & $\rm ENS$ & $\quad$ & $\rm MAD$ & $\rm OLS$ & $\rm EMD$ & $\rm SSIM$  \\
		\midrule
		IID & 0.26 & & 0.23 & 0.32 & 0.21 & 0.33 \\
		OOD & 0.25 & & 0.22 & 0.32 & 0.21 & 0.31 \\
		Extreme & 0.19 & & 0.22 & 0.28 & 0.16 & 0.16 \\
		Seasonal & 0.27 & & 0.23 & 0.38 & 0.20 & 0.32 \\
		\bottomrule
	\end{tabularx}}
	\captionof{table}{Persistence baseline performance.}
	\label{tab:baseline}
\end{wrapfigure}

\section{Challenge}\label{sec:challenge}
The EarthNet2021 challenge aims at Earth surface forecasting model intercomparison. Due to its novelty, there is not yet a commonly used criterion for Earth surface predictions.

\paragraph{EarthNetScore.}
Specifically for Earth surface forecasting, we define the \emph{EarthNetScore} as a ranking criterion balancing multiple goals in a harmonic mean as follows:
{\small
\begin{equation}\label{eq:ens}
\rm ENS = \frac{4}{(\frac{1}{\rm MAD} + \frac{1}{\rm OLS} + \frac{1}{\rm EMD} + \frac{1}{\rm SSIM})}.
\end{equation}}
The four components of $\rm ENS$ are the median absolute deviation $\rm MAD$; the difference of ordinary least squares linear regression slopes of pixelwise NDVI timeseries $\rm OLS$; the Earth mover distance $\rm EMD$ between pixelwise NDVI time series and the structural similarity index $\rm SSIM$. All component scores are modified to work properly in the presence of a data quality mask, normalized to lie between 0 (worst) and 1 (best) and rescaled to match difficulty. Since Earth surface forecasting is a stochastic task models may predict multiple future trajectories. Over a full test set, we aggregate these by only considering the best predicted trajectory per multicube (in line with video prediction common practice). Then we average the component scores of these and calculate the $\rm ENS$ by feeding the averages to eq.~\ref{eq:ens}. Thus, the $\rm ENS$ ranges from $0$ (bad) to $1$ (perfect).

\paragraph{Tracks.}
Multiple models are compared within the EarthNet2021 challenge by measuring their EarthNetScores on various tracks (see Fig.~\ref{fig:eval}). The \emph{main (IID) track} checks model validity. Models get 10 context frames of high resolution 5-daily multispectral satellite imagery (time [t-45,~t]), mesoscale dynamic climate conditions for in total 150 past and future days (time [t-50,~t+100]) and static topography at both resolutions. Models shall output 20 frames of sentinel 2 bands red, green, blue and near-infrared for the next 100 days (time [t+5,~t+100]). These predictions are evaluated with the EarthNetScore on unmasked (cloud-free) pixels from the ground truth. The \emph{Robustness (OOD) track} checks model performance on an out-of-domain (OOD) test set, since even on the same satellite data, deep learning models might generalize poorly across geolocations \citep{benson2020}. Furthermore, EarthNet2021 contains two tracks focused on Earth system science hot topics, which should both be understood as more experimental. The \emph{extreme summer track} contains cubes from the extreme summer 2018 in northern Germany \citep{bastos2020}, with 4 months of context (20 frames) starting from February and 6 months (40 frames) starting from June to evaluate predictions. The \emph{seasonal cycle track} contains multicubes with 1 year (70 frames) of context frames and 2 years (140 frames) to evaluate predictions, thus checking models applicability to the vegetation cycle.

\paragraph{EarthNet2021 Framework.}
To facilitate research we provide the EarthNet2021 framework. It contains 1) the packaged evaluation pipeline as the EarthNet2021 toolkit which leverages multiprocessing for fast inference, 2) the model intercomparison suite, which gives one entry point for running a wide range of models and 3) a naive baseline (cloud-free mean, see table~\ref{tab:baseline}) and templates for PyTorch and Tensorflow. Further information can be found on \url{www.earthnet.tech}.

\section{Outlook}
Forecasting impacts of climate and weather on the Earth surface is a simultaneously societally important and scientifically challenging task. With the EarthNet2021 dataset, first models for Europe can be designed and the EarthNet2021 challenge offers a model intercomparison framework for identifying their strengths and limitations. We expect deep learning based video prediction models to be great starting points for solutions, in perspective allowing for high-resolution prediction of localized climate impacts.

{
\paragraph{Author contributions.} CR and VB developed the dataset and challenge, wrote the manuscript and created figures. CR wrote the persistence baseline and the model intercomparison framework. VB wrote the EarthNetScore implementation and the dataset generation of EarthNet2021. JD provided Resources and helpful comments. JD, JR and MR contributed by improving the manuscript and with general discussion. MR steered and supervised, provided resources and helped with conceptual design.\\
\textbf{Acknowledgments.} We are thankful for invaluable help, comments and discussions to the DLR Climate Informatics and the MPI-BGC EIES group members, especially to Andreas Gerhardus, Christopher Käding, Miguel Mahecha, Christian Reimers, Xavier-Andoni Tibau and Rafael Vieira Westenberger. We are equally thankful to three anonymous reviewers. We estimate this project has caused around $1$ ton of carbon emissions, which we commit to offset.
}

\bibliography{refs}
\bibliographystyle{iclr2021_conference}

\end{document}